# Building Privacy-Preserving and Secure Geospatial Artificial Intelligence Foundation Models

Jinmeng Rao
University of Wisconsin-Madison, USA
jinmeng.rao@wisc.edu

Song Gao
University of Wisconsin-Madison, USA
song.gao@wisc.edu

Gengchen Mai
University of Georgia, USA
gengchen.mai25@uga.edu

Krzysztof Janowicz
University of Vienna, Austria
krzysztof.janowicz@univie.ac.at

## ABSTRACT

In recent years we have seen substantial advances in foundation models for artificial intelligence, including language, vision, and multimodal models. Recent studies have highlighted the potential of using foundation models in geospatial artificial intelligence, known as GeoAI Foundation Models, for geographic question answering, remote sensing image understanding, map generation, and location-based services, among others. However, the development and application of GeoAI foundation models can pose serious privacy and security risks, which have not been fully discussed or addressed to date. This paper introduces the potential privacy and security risks throughout the lifecycle of GeoAI foundation models and proposes a comprehensive blueprint for research directions and preventative and control strategies. Through this vision paper, we hope to draw the attention of researchers and policymakers in geospatial domains to these privacy and security risks inherent in GeoAI foundation models and advocate for the development of privacy-preserving and secure GeoAI foundation models.

## CCS CONCEPTS

• **Artificial intelligence**; • **Security and privacy**;

## KEYWORDS

GeoAI, Foundation Model, Privacy, Security, Multimodality

## 1 INTRODUCTION

Foundation Models (FMs) are large Artificial Intelligence (AI) models pre-trained on vast web-scale data and can be adapted to address a variety of downstream tasks such as machine translation and image recognition. Depending on the modalities involved, foundation models can be further categorized into language foundation models (e.g., GPT-3 [4], LLaMA [32]), vision foundation models (e.g., Segment Anything [15]), or multimodal foundation models such as vision-language foundation models (e.g., GPT-4 [24], BLIP-2 [16]) and those connecting more modalities including video and audio (e.g., ImageBind [9]). In recent years, chatbots such as ChatGPT and Bard as well as generic vision tools including Segment Anything and Stable Diffusion have showcased the proficiency of foundation models in addressing a wide range of natural language processing and computer vision tasks.

The success of foundation models has motivated researchers to incorporate them into geospatial domains to tackle challenges in Geospatial Artificial Intelligence (GeoAI) [12], known as GeoAI Foundation Models or Geo-Foundation Models (GeoFM) [21]. Recent explorations have delved into areas like geoparsing [20], urban planning [33], geographic question answering [21], remote sensing semantic segmentation [38, 8], and map generation [13], among others, yielding promising results. However, recent studies also reveal that the development and use of foundation models could potentially unveil substantial privacy and security risks, including the disclosure of sensitive information, representational bias, hallucinations, and misuse [3, 11]. These risks have sparked widespread public concerns and triggered the imposition of prohibitions and strict regulations in many countries and regions. Early regulations can be inadequate for several reasons, e.g., **1)** technology evolves faster than regulations; **2)** ensuring compliance and monitoring is challenging; and **3)** progress in domains that would benefit most can slow down. In this vision paper, we summarize the potential privacy and security risks in GeoAI foundation models and propose a blueprint for building privacy-preserving and secure GeoAI foundation models with corresponding research directions and promising preventative and control strategies. As foundation models increasingly exhibit dominance, we hope they also raise awareness of the equal importance of privacy and security, spurring more researchers to study these aspects of GeoAI foundation models.

## 2 PRIVACY AND SECURITY RISKS

Privacy and security are two intertwined concepts. Generally speaking, privacy refers to people's personal or sensitive information and their rights to prevent the disclosure of such information. Security refers to how such information is protected. In geospatial domains, privacy and security often concern sensitive geospatial information such as home location, workspace, Points-of-Interest (POI) preferences, daily trajectories, and inferences based on such information [14, 28]. In the lifecycle of building and utilizing GeoAI foundation models, we identify a series of potential privacy and security risks that exist around the pre-training and fine-tuning stages with geospatial data, centralized serving and tooling, prompting-based interaction, and feedback mechanisms.

### 2.1 Risks in Geospatial Pre-training

As large pre-trained models, the capability of foundation models relies heavily on the scale, quality, and diversity of their pre-training data. For example, GPT-3 was pre-trained on a huge language corpus consisting of around 500 billion tokens from Web resources (e.g., CommonCrawl, Wikipedia) and books. BLIP-2 and DINOv2 were pre-trained on 129 million images (with captions) and 142 million images, respectively. Similarly, to pre-train a GeoAI foundation model, large-scale geospatial data are the key. Given the multimodal nature of geospatial domains, geospatial data involve various modalities such as language (e.g., street address, geo-tagged social media posts), vision (e.g., remote sensing and street view imagery, atlases), and structured data such as vectors (e.g., trajectories), graphs (e.g., geospatial knowledge graphs), and tabular data (e.g., census). All of them may contain personal or sensitive geospatial information that can be learned and disclosed by GeoAI foundation models. For example, a language model could potentially memorize home addresses from a pre-training corpus and disclose them to anyone who asks. A vision-language model, likewise, may learn the alignment between a building and its residents who have mentioned it on social media. If someone were to upload a picture of that building and asked, *"Who lives in this building?"*, the model might list all the residents it recognizes from social media. Since how foundation models determine to learn and utilize the data remains opaque, it is challenging to prevent a GeoAI foundation model from acquiring, retaining, and divulging sensitive geospatial information without adequate privacy and security measures.

## 2.2 Risks in Geospatial Fine-tuning

In practice, due to limitations in computational resources and the availability of domain-specific data, we usually adopt model weights from foundation models pre-trained on a general domain and further fine-tune it on domain-specific data (i.e., domain adaptation) such as geospatial data. Common fine-tuning methods for foundation models include model fine-tuning, where models are fine-tuned on domain-specific data (for instance, instruction tuning focuses on fine-tuning models to follow given instructions), and prompt tuning, which involves fine-tuning input prompts rather than the models themselves. There are several risks associated with the fine-tuning data and the fine-tuning process: **1) Memorizing sensitive data.** Analogous to geospatial pre-training, if the fine-tuning data contains personal or sensitive geospatial information, a GeoAI foundation model may potentially learn, memorize, and disseminate such information after the model has been fine-tuned; **2) Poisoning by malicious instructions.** During instruction tuning, if the instruction dataset has been poisoned or injected with malicious instructions such as backdoors [34], the fine-tuned models can be easily manipulated and perform malicious activities (e.g., analyzing and revealing home locations of individuals); and **3) Attacks due to leaked soft prompts.** For prompt tuning, since the soft prompts (e.g., embeddings) are tuned on user input data, once such soft prompts are leaked, attackers might be able to infer user input information (e.g., frequently mentioned places) from these prompts.

## 2.3 Risks in Centralized Serving and Tooling

After training, GeoAI foundation models are usually hosted on centralized servers. There are two paradigms of how GeoAI foundation models can be used to provide services. One paradigm is that we fully rely on the internal knowledge of models learned during the pre-training or fine-tuning stage to provide services such as question answering. Another paradigm is that we enable the models with geospatial tooling ability by connecting the models to external geospatial resources such as geospatial databases, tools, and APIs via autonomous LLM frameworks [17, 7] such as LangChain [1] and AutoGPT [2] so that the models can acquire and utilize external geospatial knowledge on which they were not trained to further enhance their functionality. Both paradigms may expose privacy and security risks. First, centralized serving brings endogenous privacy risks as all users' requests need to be sent to and stored in a centralized server. Sometimes, these requests may contain sensitive geospatial information that users are unwilling to disclose. Recent payment leakage [3] and chat history leakage [4] in ChatGPT show that as soon as the data leaves a user's device, avoiding privacy and security risks becomes very challenging. Second, since the model weights are stored in a centralized server, an attacker may be able to hack into the server to steal the weights, reconstruct training data from the weights [10], or perform membership inference attacks [23]. Third, when GeoAI foundation models are connected to external geospatial resources, attackers might make models disclose sensitive information from external resources (e.g., geospatial database credentials or third-party private geospatial data).

## 2.4 Risks in Prompt-Based Interaction

Prompt-based interactions are widely supported in most foundation models. Properly designed prompts can leverage the in-context learning (e.g., zero-shot or few-shot learning) ability of foundation models to tackle a wide variety of tasks. Many geospatial applications built upon foundation models are directly based on prompt engineering. For example, ChatGeoPT [5] and MapsGPT [6] both incorporate language foundation models with pre-defined prompt templates covering the use of Location-Based Service (LBS) APIs to provide users with flexible location search experience via natural language. In contrast, improper prompts can be used to "jailbreak" models to get sensitive information or "hijack" models to perform something dangerous, illegal, or unethical. Common prompt attack methods include Do Anything Now [7] (i.e., bypass pre-set content policy), Goal Hijacking [25] (i.e., ignore pre-defined prompts), etc., which can cause serious privacy and security issues: **1) Training data leakage.** Recent studies [5, 36] reveal that by constructing certain types of prompts or prefixes, attackers can make foundation models output the content they memorized during the training stage, such as home locations of individuals; **2) Pre-defined prompt leakage.** Attackers may use certain prompts to ask the models to divulge their pre-defined prompts, which may contain sensitive information such as instructions to access internal geospatial systems. Recent prompt leakage accidents of Bing Chat and Snap's MyAI suggest that it is hard to make foundation models fully immune to such prompt attacks even with privacy and security measures established; **3) External resource leakage.** As mentioned previously, attackers may use malicious prompts to induce models to send queries to the connected geospatial database and get sensitive geospatial information; and **4) Malicious behaviors.** Attackers may use malicious prompts to induce models to send phishing emails, scams, or commit cybercrimes [6].

---

[1] https://python.langchain.com/
[2] https://github.com/Significant-Gravitas/Auto-GPT
[3] https://cybernews.com/news/payment-info-leaked-openai-chatgpt-outage/
[4] https://www.bbc.com/news/technology-65047304
[5] https://github.com/earth-genome/ChatGeoPT
[6] https://www.mapsgpt.com/
[7] https://github.com/0xk1h0/ChatGPT_DAN

## 2.5 Risks in Feedback Mechanisms

Feedback mechanisms play an important role in continuously improving the quality of foundation models and making foundation models helpful and safe. Two common paradigms are Reinforcement Learning from Human Feedback (RLHF) [1] and Reinforcement Learning from AI Feedback (RLAIF) [2]. The former utilizes human evaluation as rewards or penalties to improve models, while the latter asks AI to improve itself. In some cases, these feedback mechanisms can also be utilized by attackers to impact model quality and even make models toxic. For example, attackers can create plenty of misleading feedback or conduct backdoor reward poisoning attack [39, 31] against the RL process, causing the models to produce false geographical knowledge, geographical discrimination remarks, or controversial geopolitical statements.

## 3 TOWARDS PRIVACY-PRESERVING AND SECURE GEOAI FOUNDATION MODELS

Motivated by the above-mentioned privacy and security risks, we advocate for building privacy-preserving and secure GeoAI foundation models. We propose a blueprint (shown in Figure 1) that outlines multiple research directions, along with their associated challenges and promising approaches.

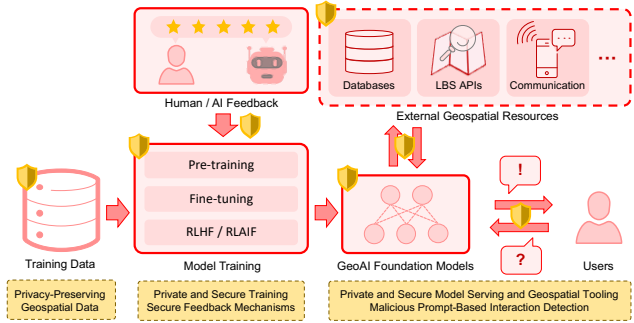

**Figure 1: A blueprint of building privacy-preserving and secure GeoAI foundation models.**

## 3.1 Privacy-Preserving Geospatial Data

Large-scale privacy-preserving geospatial data is key to building privacy-preserving and secure GeoAI foundation models. The multimodal nature of geospatial data implies that different types of geospatial data require different privacy protection methods (e.g., geomasking, K-anonymity, and differential privacy for locations and trajectories, and blurring and mosaic for street view images), and some privacy risks exist in a cross-modal fashion (e.g., revealing home locations by associating social media posts and street view images). Thus, three challenges arise in this direction: **1)** how to effectively process and preserve the privacy of large-scale multimodal geospatial data. This is a novel challenge to GeoAI foundation models as we must address cross-modal privacy risks in geospatial data while maintaining geospatial alignment; **2)** how to strike a balance between geospatial data privacy and utility. This becomes more complex in GeoAI foundation models compared to traditional approaches as the dimensions of modalities, levels of scales, and data size increase significantly; and **3)** how to measure and mitigate potential geographic biases [18] existing in large-scale multimodal geospatial data. Promising approaches include using spatial cluster computing frameworks such as Apache Sedona for large-scale geospatial data processing, using deep learning methods to protect location privacy and balance the privacy-utility trade-off [28, 26], using multimodal instruction tuning to enhance alignment among modalities and reduce biases and hallucinations [19], etc.

## 3.2 Private and Secure Training and Serving

Traditional centralized training and serving strategies require training data and model weights to be stored on a centralized server. However, in most cases, users are reluctant to share their data due to privacy concerns. This inherent risk in the centralized structure poses three challenges for GeoAI foundation models: **1)** how to conduct geospatial pre-training and fine-tuning in ways that ensure both privacy and security; **2)** how to ensure the privacy and security of hosted GeoAI foundation models (e.g., preventing model weight leakage and inference attacks); and **3)** when switching to decentralized training and serving strategies, how to deal with non-Independently Identically Distributed (non-IID) training data. This scenario frequently occurs when data is collected across different geographic regions due to spatial heterogeneity. Promising approaches include efficient encoding and encryption for geospatial data, model weights, and data transfer procedures, federated learning for pre-training, fine-tuning, and prompt engineering [27, 37, 35], geospatial-aware contrastive pre-training [22], etc.

## 3.3 Private and Secure Geospatial Tooling

Geospatial tooling greatly enhances the usability and extensibility of GeoAI foundation models. For example, geographic information retrieval can improve the truthfulness and timeliness of the models, and various geospatial resources and services can equip the models with location tracking, spatial analysis, and geocomputing capabilities. However, allowing GeoAI foundation models to connect to various geospatial tools without restrictions might result in severe privacy and security issues such as sensitive geospatial data leakage and misuse. This raises two challenges: **1)** how to design a generic and secure protocol to regulate geospatial tooling for GeoAI foundation models; and **2)** how to teach models to use geospatial tools and interpret results in ways that uphold privacy and security. Promising approaches include adopting in-context learning [30], fine-tuning [29, 1, 2], and autonomous agents [17] to improve and regulate geospatial tooling for GeoAI foundation models.

## 3.4 Private and Secure Interaction with GeoFMs

Prompt-based interaction facilitates natural communication between users and GeoAI foundation models, but it also raises privacy and security concerns. The key to ensuring privacy-preserving and secure prompt-based interaction is to identify malicious prompts from the myriad of user inputs. This presents three challenges: **1)** how to understand the intentions behind prompts and filter out malicious prompts; **2)** how to evaluate the resilience and robustness of GeoAI foundation models against malicious prompts; and **3)** how to ensure users do not accidentally send sensitive geospatial information to GeoAI foundation models via prompts and vice versa. Promising approaches include developing a generic detector that can detect or filter out either malicious prompts or sensitive geospatial information from the users' side, establishing an evaluation framework covering different types of malicious prompts to test GeoAI foundation models' resilience and robustness, etc.

## 3.5 Secure Feedback Mechanisms

Feedback mechanisms such as RLHF and RLAIF can be exploited by attackers using poisoned feedback, leading to harmful GeoAI foundation models that produce inaccurate, controversial, or unethical geographic statements. In terms of ensuring secure feedback mechanisms, we highlight two challenges: **1)** how to identify poisoned feedback and determine if, when, where, and how the feedback mechanisms or reward functions are compromised (e.g., poisoned feedback could be comments or ratings that induce the model to deviate from expected behavior); and **2)** how to determine the best recovery or fallback strategy when we realize the feedback mechanisms are poisoned. In these cases, we can evaluate the model's behavior using a comprehensive benchmark set of geographic commonsense knowledge and conversations and measure the shift of truthfulness and harmfulness between versions. In addition, analyzing the data distribution of feedback, monitoring the reward curve, and regularly saving model weights are also beneficial.

## 4 CONCLUSION

This vision paper discusses privacy and security risks in GeoAI foundation models and proposes a blueprint for privacy-preserving and secure GeoAI foundation models. Addressing privacy and security risks in new technologies requires not only regulations but also evolving technical solutions and societal ethical discussions. We hope this paper can raise awareness among researchers and policymakers about the privacy and security risks associated with GeoAI foundation models and promote positive future development of GeoAI foundation models that respect privacy and security.

## ACKNOWLEDGMENTS

Song Gao acknowledges the funding support from the National Science Foundation funded AI institute [Grant No. 2112606] for Intelligent Cyberinfrastructure with Computational Learning in the Environment (ICICLE).